\documentclass[10pt,twocolumn,letterpaper]{article}

\usepackage{wacv}              % To produce the CAMERA-READY version
\usepackage[accsupp]{axessibility} 
\definecolor{wacvblue}{rgb}{0.21,0.49,0.74}
\usepackage[pagebackref,breaklinks,colorlinks,allcolors=wacvblue]{hyperref}

\def\wacvPaperID{3493} % *** Enter the WACV Paper ID here
\def\confName{WACV}
\def\confYear{2027}

\title{PixelSpline: Continuous Space-Time Video Super-Resolution via Endpoint-Anchored Shared Pixel Trajectories}

\author{
Wanqi Yuan \qquad Nianyi Li\\
Clemson University\\
Clemson, South Carolina, United States\\
{\tt\small wanqiy@clemson.edu \qquad nianyil@clemson.edu}
}

\begin{document}
\maketitle
\begin{abstract}
Existing continuous space-time video super-resolution methods often infer motion from only two endpoint frames and model the two motion directions independently, making them vulnerable to ambiguous correspondence and inconsistent trajectories. We introduce a continuous STVSR framework that combines long-context encoding, continuous HR feature lifting, and an endpoint-anchored PixelSpline. Long-range observations provide more reliable endpoint representations, while continuous upsampling constructs detailed appearance features directly on the requested HR grid. PixelSpline explicitly models each pixel's motion as a single continuous trajectory shared by both endpoints. Using this trajectory, our method warps endpoint colors into a coarse intermediate frame and aligns the HR features for residual reconstruction. Experiments across spatial and temporal upsampling settings demonstrate that the proposed framework offers a favorable quality--efficiency trade-off. Our project page is: \url{https://wanqiyuan.github.io/PixelSpline/}
\end{abstract}    
\section{Introduction}
\begin{figure}[t]
    \centering
    \includegraphics[width=\columnwidth]
        {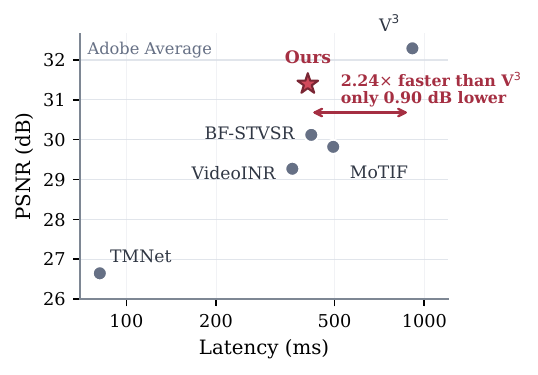}
    \caption{
        Quality--latency trade-off on Adobe240 Average under $\mathrm{S}\!\times\!4$ spatial and $\mathrm{T}\!\times\!8$ temporal upsampling. Our method is $2.24\times$ faster than V$^3$, with a $0.90$\,dB PSNR difference. Latency is measured end-to-end on a single NVIDIA H100 GPU after warm-up.
    }
    \label{fig:quality_latency_tradeoff}
\end{figure}
Video super-resolution (VSR) aims to recover high-resolution (HR) videos from low-resolution (LR) observations. It is important for video capture, display, editing, and analysis, where spatial resolution is often limited by camera hardware, transmission bandwidth, or storage. However, improving spatial resolution alone does not resolve temporal undersampling: a sharp video can still exhibit discontinuous motion when recorded at a low frame rate. Video frame interpolation (VFI) addresses this limitation by synthesizing frames at unobserved time steps, but it cannot recover spatial details already lost in the LR inputs. Space-time video super-resolution (STVSR) therefore combines VSR and VFI to reconstruct videos with both higher spatial resolution and higher frame rate. More recently, continuous STVSR (C-STVSR) has generalized this objective further, allowing a video to be queried at arbitrary spatial scales and continuous time coordinates.

% Early STVSR approaches either cascade VFI and VSR models or jointly optimize the two tasks for predefined integer upsampling factors. 
Early STVSR approaches either connect VFI and VSR models with two-stage settings or jointly optimize the two tasks for predefined integer upsampling factors. One-stage methods such as Zooming Slow-Mo~\cite{xiang2020zooming} and TMNet~\cite{xu2021temporal} demonstrate the benefit of joint spatial and temporal reconstruction. Continuous methods instead model video as a function over space and time. VideoINR~\cite{chen2022videoinr} uses implicit functions for spatial appearance and time-dependent motion. MoTIF~\cite{chen2023motif} models forward motion trajectories and splats endpoint features to a queried time, while BF-STVSR~\cite{kim2025bf} introduces B-spline motion modeling together with a Fourier-based spatial representation. V$^{3}$~\cite{becker2025continuous} recently explores a unified 3D Fourier field that avoids explicit warping.
% Despite their different formulations, most existing C-STVSR systems either
% construct an intermediate frame primarily from the two adjacent endpoints or
% encode motion and appearance in a representation whose geometric behavior is
% not directly constrained by observations beyond that pair.

Despite this progress, three closely related challenges remain. First, a correspondence inferred from only two endpoint frames can be ambiguous in occluded or disoccluded regions, and can be corrupted by motion blur, repetitive texture, weak texture, or large displacement. A mistaken endpoint match then affects every intermediate time queried between the pair, often producing ghosting, temporal jumps, or unstable structures. A longer temporal context can reduce this ambiguity: content hidden at one endpoint may be visible in an earlier or later frame, repeated observations provide additional evidence for motion direction, and trajectories observed over several frames help distinguish persistent motion from pairwise matching noise. Second, existing continuous methods generally evaluate the motion fields from the two endpoints to the target time separately. Even when the two directions share network parameters, their predictions are conditioned on different endpoint features or motion latents, without an explicit constraint that corresponding endpoint pixels follow the same physical curve. The two warps can therefore transport corresponding content toward inconsistent target locations. Third, arbitrary-scale temporal synthesis requires endpoint appearance features on the requested HR grid. Fixed-scale upsamplers cannot serve arbitrary spatial queries, while warping only LR features postpones the construction of spatial detail until the final decoder. It is therefore desirable to lift the endpoint features continuously before temporal transport.

To address these challenges, we propose a continuous STVSR framework composed of four stages. First, a long-context encoder aggregates information across multiple LR observations to produce target-pair-aware endpoint representations, improving correspondence estimation in ambiguous regions. Second, a CUF-based continuous feature upsampling module transforms the two LR endpoint representations into HR appearance features on the requested output grid, providing spatially detailed evidence before temporal synthesis. Third, we establish dense bidirectional endpoint correspondences and use them to construct an endpoint-constrained PixelSpline for every source pixel. Rather than predicting two endpoint-to-target trajectories, we construct one curve on the first endpoint's correspondence grid and use the endpoint correspondence to express the same curve from the other endpoint. The two endpoint warps therefore traverse one shared path from opposite ends. At any target time, this trajectory is used twice: bicubic-upsampled endpoint colors are forward-warped and fused into a coarse, potentially blurry HR intermediate frame, while the lifted endpoint features are transported to the same target locations to form an aligned HR feature representation. Finally, a lightweight pointwise decoder converts the aligned representation into an HR residual, which is added to the coarse frame to obtain the final reconstruction. This design integrates geometry and appearance with distinct roles: the shared PixelSpline models how observed content moves, whereas the decoder restores details that cannot be recovered through direct pixel transport.

Extensive experiments on standard STVSR benchmarks show that the proposed framework occupies a favorable point on the quality--efficiency trade-off. As shown in Fig.~\ref{fig:quality_latency_tradeoff}, our method runs $2.24\times$ faster than the quality-leading V$^3$ baseline, with a PSNR that is only 0.90 dB lower.

Our main contributions are summarized as follows:
\begin{itemize}
    \item We introduce long-context endpoint encoding that uses observations beyond the queried pair to obtain more stable and globally informed endpoint representations.

    \item We propose an endpoint-anchored shared PixelSpline that explicitly models per-pixel motion as one continuous trajectory evaluated from both endpoints.

    \item We develop a CUF-based continuous feature lifting module that adapts continuous upsampling filters to map context-enhanced LR endpoint features onto arbitrary HR grids before trajectory-guided temporal transport.

\end{itemize}

\section{Related Work}
\paragraph{Arbitrary-scale super-resolution.}
% Early arbitrary-scale image super-resolution predicts scale-conditioned
% reconstruction filters, as in Meta-SR~\cite{hu2019meta}. Coordinate-based
% methods subsequently model an image as a continuous function: LIIF~\cite{chen2021learning}
% decodes local features at arbitrary queries, while LTE~\cite{lee2022local} and
% Thera~\cite{becker2023thera} use sinusoidal representations to better preserve
% high-frequency content and control aliasing across scales. SAVSR~\cite{li2024savsr}
% and ST-AVSR~\cite{shang2024arbitrary} extend arbitrary spatial magnification to
% video by exploiting neighboring frames, but retain the original temporal grid.
% These developments motivate lifting video features continuously in space, but
% do not by themselves address interpolation at unseen times.
Early arbitrary-scale image super-resolution predicts scale-conditioned reconstruction filters, as in Meta-SR~\cite{hu2019meta}. Coordinate-based methods subsequently model an image as a continuous function: LIIF~\cite{chen2021learning} decodes local features at arbitrary queries, while LTE~\cite{lee2022local} and Thera~\cite{becker2023thera} use sinusoidal representations to better preserve high-frequency content and control aliasing across scales. SAVSR~\cite{li2024savsr} and ST-AVSR~\cite{shang2024arbitrary} extend arbitrary spatial magnification to video by exploiting neighboring frames, but retain the original temporal grid. These developments motivate lifting video features continuously in space, but do not address interpolation at unseen times by themselves.

\paragraph{Space-time video super-resolution.}
% Classical STVSR pipelines combine frame interpolation~\cite{jiang2018super,bao2019depth}
% with image or video super-resolution~\cite{zhang2018image,wang2019edvr}.
% One-stage methods instead learn spatial and temporal reconstruction jointly.
% Zooming Slow-Mo~\cite{xiang2020zooming} employs bidirectional deformable
% ConvLSTMs, TMNet~\cite{xu2021temporal} introduces temporal modulation, and
% subsequent architectures improve cross-frame propagation and aggregation
% ~\cite{haris2020space,geng2022rstt,hu2022you,hu2023store}. These models are
% effective at their trained integer factors, but do not provide continuous
% space-time queries. Recently, Time Without Time~\cite{choi2026time} improves their
% pretraining through masked pseudo-temporal sequences, while retaining the same
% fixed-scale formulation.
Classical STVSR pipelines combine frame interpolation~\cite{jiang2018super,bao2019depth} with image or video super-resolution~\cite{zhang2018image,wang2019edvr}. One-stage methods instead learn spatial and temporal reconstruction jointly. Zooming Slow-Mo~\cite{xiang2020zooming} employs bidirectional deformable ConvLSTMs, TMNet~\cite{xu2021temporal} introduces temporal modulation, and subsequent architectures improve cross-frame propagation and aggregation ~\cite{haris2020space,geng2022rstt,hu2022you,hu2023store}. These models are effective at their trained integer factors, but do not provide continuous space-time queries. Recently, Time Without Time~\cite{choi2026time} improves their pretraining through masked pseudo-temporal sequences, while retaining the same fixed-scale formulation.

% Continuous STVSR removes this restriction by treating space and time as query
% coordinates. VideoINR~\cite{chen2022videoinr} combines an implicit spatial
% representation with backward motion, and MoTIF~\cite{chen2023motif} models
% forward trajectories and splats endpoint features to the target time.
% BF-STVSR~\cite{kim2025bf} augments this formulation with Fourier features and
% B-spline motion, whereas V$^3$~\cite{becker2025continuous} represents the video
% as a unified 3D Fourier field without explicit warping. Pairwise trajectory
% methods, however, construct a query from two anchor-attached representations
% and model the two transport directions separately; V$^3$ uses broader temporal
% context but does not impose explicit correspondence-consistent geometry. Our
% method brings these complementary ideas together by using long-context endpoint
% features to construct one endpoint-anchored PixelSpline shared by both
% transport directions.
Continuous STVSR removes this restriction by treating space and time as query coordinates. VideoINR~\cite{chen2022videoinr} combines an implicit spatial representation with backward motion, and MoTIF~\cite{chen2023motif} models forward trajectories and splats endpoint features to the target time. BF-STVSR~\cite{kim2025bf} augments this formulation with Fourier features and B-spline motion, whereas V$^3$~\cite{becker2025continuous} represents the video as a unified 3D Fourier field without explicit warping. Pairwise trajectory methods, however, construct a query from two anchor-attached representations and model the two transport directions separately; V$^3$ uses broader temporal context but does not ensure explicit correspondence-consistent geometry. Our method brings these complementary ideas together by using long-context endpoint features to construct one endpoint-anchored PixelSpline shared by both transport directions.

\section{Method}

\subsection{Overview}
\begin{figure*}[t]
    \centering
    \includegraphics[width=\textwidth]{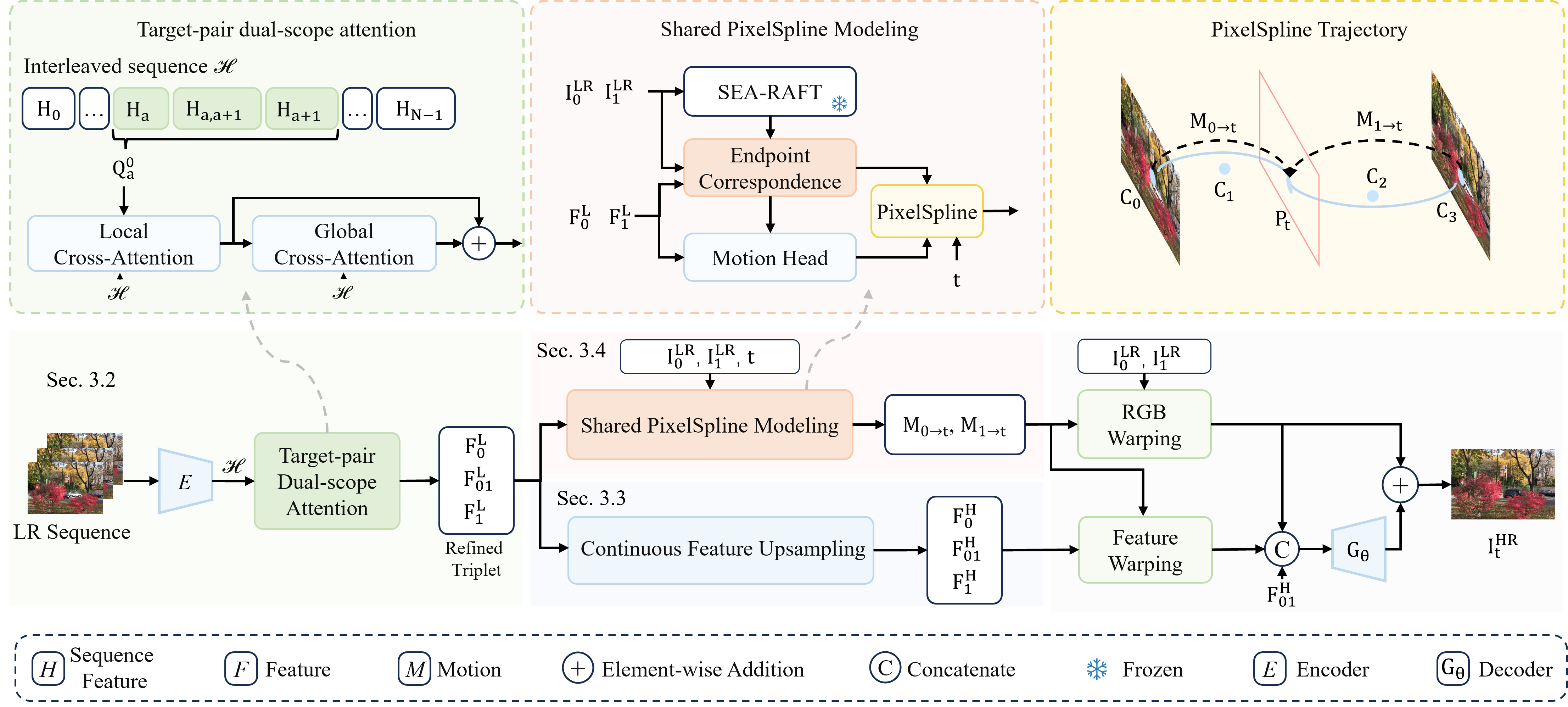}
    \caption{ 
%     Overview of our continuous STVSR framework. A bidirectional long-context
% encoder aggregates information from multiple LR frames to obtain
% context-enhanced endpoint representations \textbf{(Sec.3.2)}. Continuous feature upsampling maps the endpoint features onto the requested HR grid \textbf{(Sec.3.3)}, while an endpoint-anchored PixelSpline constructs a shared trajectory for RGB and feature transport at the queried time \textbf{(Sec. 3.4)}. Finally, a lightweight decoder predicts an HR residual that is added to the coarse warped frame to produce the reconstruction.
Overview of our continuous STVSR framework. A bidirectional long-context encoder aggregates information from multiple LR frames to obtain context-enhanced endpoint representations \textbf{(Sec.3.2)}. Continuous feature upsampling maps the endpoint features onto the requested HR grid \textbf{(Sec.3.3)}, while an endpoint-anchored PixelSpline constructs a shared trajectory for RGB and feature transport at the queried time \textbf{(Sec. 3.4)}. Finally, a lightweight decoder predicts an HR residual that is added to the coarse warped frame to produce the reconstruction. 
% For clarity, auxiliary rendering cues and decoder conditioning are not shown in detail.
}
    \label{fig:pipeline}
\end{figure*}

Let $\mathcal{I}^{L}=\{I_n^{L}\}_{n=1}^{N}$ be a window of LR frames, and let $I_0^{L}$ and $I_1^{L}$ be the two anchor frames surrounding a query time $t\in[0,1]$. Given an arbitrary spatial scale $s$, our goal is to reconstruct the HR intermediate frame $I_t^{H}$.

Our framework contains four main components. First, a long-context encoder uses all frames in $\mathcal{I}^{L}$ to produce context-enhanced features for the two anchors. Second, a continuous feature lifting module maps these LR anchor features to the requested HR grid. Third, PixelSpline constructs one explicit trajectory for every matched point from the first anchor to the second. At any query time, the trajectory has a unique position on the intermediate frame. We use this position to transport both anchor pixels and their lifted HR features, producing a coarse intermediate frame and an aligned HR feature. Finally, a lightweight decoder predicts an image residual from the aligned feature and adds it to the coarse frame.

We denote the encoder output by
\begin{equation}
    (\mathbf{F}_0^{L},\mathbf{F}_{01}^{L},\mathbf{F}_1^{L})
    =E(\mathcal{I}^{L}),
    \label{eq:encoder_overview}
\end{equation}
where $\mathbf{F}_0^{L}$ and $\mathbf{F}_1^{L}$ are the two anchor features, and $\mathbf{F}_{01}^{L}$ describes their shared interval. The upsampling module produces $\mathbf{F}_0^{H}$, $\mathbf{F}_1^{H}$ and $\mathbf{F}_{01}^{H}$ at scale $s$. PixelSpline then transports the anchor RGB values and HR features to time $t$, yielding a coarse frame $\widetilde I_t^{H}$ and an aligned feature
$\mathbf{F}_t^{H}$. The final result is
\begin{equation}
    \widehat I_t^{H}
    =\widetilde I_t^{H}
    +G_{\theta}(\mathbf{F}_t^{H}, \mathbf{F}_{01}^{H}, \widetilde I_t^{H}, t,s),
    \label{eq:final_output}
\end{equation}
where $G_{\theta}$ is the residual decoder. Fig.~\ref{fig:pipeline} illustrates the pipeline of our method.

\subsection{Target-Pair-Aware Context Encoding}
\label{sec:long_context}

We adopt the native multi-frame encoder of Zooming Slow-Mo~\cite{xiang2020zooming} to extract temporally aggregated features from the LR input sequence. We augment its output with a target-pair attention module. This module enables the two anchor frames to exchange features with relevant frames across the entire encoded sequence.

\paragraph{Multi-frame sequence encoding.}
Given $N$ LR frames, the encoder first extracts a three-level feature pyramid for each frame, we then align the features of adjacent frames. The two aligned features are fused by a $1\times1$ convolution to obtain an interval feature. The original frame features and the interval features are arranged into an interleaved sequence,
\begin{equation}
    \mathcal{H}
    =\{\mathbf{H}_0,\mathbf{H}_{0,1},\mathbf{H}_1,
       \mathbf{H}_{1,2},\ldots,\mathbf{H}_{N-1}\},
    \label{eq:interleaved_features}
\end{equation}
which contains $2N-1$ elements. A bidirectional deformable ConvLSTM propagates information through this sequence in both temporal directions. Consequently, every element in $\mathcal{H}$ contains multi-frame context while retaining its role as either a frame feature or an interval feature.

\paragraph{Target-pair dual-scope attention.}
The multi-frame encoder treats all temporal positions uniformly, whereas our subsequent reconstruction concerns one particular interval. Let $a$ denote the position of the queried pair in the input window. We take its left-anchor, interval, and right-anchor features as the initial query,
\begin{equation}
    \mathbf{Q}_a^{0}
    =\big[\mathbf{H}_{a},\mathbf{H}_{a,a+1},
           \mathbf{H}_{a+1}\big],
    \label{eq:target_triplet}
\end{equation}
while the complete sequence $\mathcal{H}$ serves as key--value memory.

We refine this triplet at both local and global spatial scopes. The local branch applies cross-attention within regular non-overlapping windows and then within shifted windows. The first pass retrieves spatially corresponding evidence from all temporal positions, while the shifted pass enables information exchange across window boundaries. The global branch pools every feature map to a compact spatial grid and performs cross-attention over the pooled query and memory tokens. Its output is upsampled and added to the local features, providing scene-level context at a low computational cost. We also add temporal-offset embeddings to identify the distance of each memory element from the queried interval, and type embeddings to distinguish frame features from interval features.

The attention module returns the refined target triplet
\begin{equation}
    (\mathbf{F}_0^{L},\mathbf{F}_{01}^{L},\mathbf{F}_1^{L})
    =A_{\mathrm{ctx}}(\mathbf{Q}_a^{0},\mathcal{H}),
    \label{eq:context_attention}
\end{equation}
where $\mathbf{F}_0^{L}$ and $\mathbf{F}_1^{L}$ are the context-enhanced anchor features and $\mathbf{F}_{01}^{L}$ represents their interval. 
% The two anchor features are passed to the continuous lifting and motion modules, while the interval feature provides additional context for motion estimation.

\subsection{Continuous Feature Upsampling}

Our lifting module follows continuous upsampling filters (CUF) and the hyper-upsampling design used in ST-AVSR~\cite{shang2024arbitrary}. We adapt these image-reconstruction modules to our feature-transport setting: instead of producing RGB values, the module maps a $C$-channel LR anchor feature to a $C$-channel HR feature that will subsequently be warped by PixelSpline.

Given an LR anchor feature $\mathbf{F}_i^{L}$, $i\in\{0,1\}$, and an HR query coordinate $\mathbf{x}$, we first obtain a bilinearly upsampled base feature
\begin{equation}
    \mathbf{B}_i^{H}(\mathbf{x})
    =\operatorname{Bilinear}(\mathbf{F}_i^{L},\mathbf{x}).
    \label{eq:lifting_base}
\end{equation}
The second branch predicts a continuous, depth-wise $3\times3$ filter for each HR query. Let $\mathbf{x}_r$ be the nearest LR feature coordinate, $\Delta\mathbf{x}=\mathbf{x}-\mathbf{x}_r$ its relative offset, $\mathbf{a}$ the output cell size, and $\mathbf{o}_k$ the spatial index of the $k$-th location in a $3\times3$ kernel. After sinusoidal encoding, these three geometric quantities are passed to a SIREN-based\cite{sitzmann2020implicit} kernel predictor:
\begin{equation}
    \mathbf{w}_k(\mathbf{x},\mathbf{a})
    =K_{\phi}\!\left(
       \gamma_r(\Delta\mathbf{x}),
       \gamma_a(\mathbf{a}),
       \gamma_k(\mathbf{o}_k)
     \right), \qquad k=1,\ldots,9,
    \label{eq:cuf_kernel}
\end{equation}
where $\gamma_r$, $\gamma_a$, and $\gamma_k$ denote the three sinusoidal encoders. Each $\mathbf{w}_k\in\mathbb{R}^{C}$ contains one weight per feature channel. The filter depends only on the query geometry and output scale, not on the input feature values. Thus, the same filters can be shared by both anchors when they are lifted to the same target grid.

To apply the predicted filters, we unfold each LR feature into its $3\times3$ neighborhoods and bilinearly sample the unfolded tensor at all HR queries. Writing the resulting tensor as $\overline{\mathcal{U}}_i$ and the collection of predicted filters as $\mathbf{W}$, the CUF reconstruction is
\begin{equation}
    \mathbf{R}_i^{H}
    =\operatorname{Fold}\!\left(
       \mathbf{W}\odot\overline{\mathcal{U}}_i
     \right).
    \label{eq:cuf_reconstruction}
\end{equation}
% A $1\times1$ convolution, a LeakyReLU, and a $3\times3$ convolution transform $\mathbf{R}_i^{H}$ into a feature residual. 
$H_{\psi}$ transforms $\mathbf{R}_i^{H}$ into a feature residual with a $1\times1$ convolution, a LeakyReLU, and a $3\times3$ convolution.
The final lifted feature is
\begin{equation}
    \mathbf{F}_i^{H}
    =\mathbf{B}_i^{H}+H_{\psi}(\mathbf{R}_i^{H}).
    \label{eq:lifted_feature}
\end{equation}
This residual formulation retains the bilinearly lifted anchor content while allowing the continuous filters to recover scale-dependent feature detail. The same lifting module is applied to both anchors.

\subsection{Shared PixelSpline Modeling and Rendering}

\paragraph{Endpoint correspondence.}
Inspired by SGM-VFI\cite{liu2024sparse} and GMFlow\cite{xu2022gmflow} We first estimate a dense correspondence between the two anchor frames. A local flow estimator (SEA-RAFT\cite{wang2024sea} in our pipeline) supplies an initial match and its confidence. To correct unreliable regions, we measure the reconstruction error after warping one anchor to the other and perform global matching with frozen GMFlow at locations with large errors. The local and global candidates are merged and refined using the context-enhanced anchor features. This produces bidirectional endpoint flows $\mathbf{M}_{0\rightarrow1}$ and $\mathbf{M}_{1\rightarrow0}$ together with
their confidence maps. These endpoint correspondences determine which two pixels should be connected by PixelSpline.

\paragraph{A unique trajectory between two matched pixels.}
Consider a pixel at position $\mathbf{p}$ in the first anchor and its matched position
\begin{equation}
    \mathbf{q}=\mathbf{p}+\mathbf{M}_{0\rightarrow1}(\mathbf{p})
    \label{eq:endpoint_match}
\end{equation}
in the second anchor. Instead of predicting two independent motions from $\mathbf{p}$ and $\mathbf{q}$ to the target frame, we explicitly connect them with one cubic curve. The motion head predicts two offsets $\boldsymbol{\delta}_1$ and $\boldsymbol{\delta}_2$ from the anchor features.
We define four control points
\begin{equation}
\begin{aligned}
    \mathbf{C}_0 &= \mathbf{p},
    &\mathbf{C}_1 &= \mathbf{p}+\tfrac{1}{3}(\mathbf{q}-\mathbf{p})
                         +\boldsymbol{\delta}_1,\\
    \mathbf{C}_2 &= \mathbf{p}+\tfrac{2}{3}(\mathbf{q}-\mathbf{p})
                         +\boldsymbol{\delta}_2,
    &\mathbf{C}_3 &= \mathbf{q}.
\end{aligned}
\label{eq:absolute_controls}
\end{equation}
The resulting clamped cubic B-spline is equivalently a cubic B\'ezier segment:
\begin{equation}
    \mathbf{P}_{\mathbf{p}}(t)
    =\sum_{k=0}^{3}{3\choose k}(1-t)^{3-k}t^k\mathbf{C}_k.
    \label{eq:shared_curve}
\end{equation}
Because $\mathbf{C}_0=\mathbf{p}$ and $\mathbf{C}_3=\mathbf{q}$, the curve passes through the two matched endpoints exactly.

The geometric meaning of Eq.~\eqref{eq:shared_curve} is simple: the time-$t$ frame can be viewed as a plane that intersects the trajectory at one point, $\mathbf{p}_t=\mathbf{P}_{\mathbf{p}}(t)$. We therefore obtain the two forward motions by measuring displacement to the same intersection:
\begin{equation}
    \mathbf{M}_{0\rightarrow t}(\mathbf{p})=\mathbf{p}_t-\mathbf{p},
    \qquad
    \mathbf{M}_{1\rightarrow t}(\mathbf{q})=\mathbf{p}_t-\mathbf{q}.
    \label{eq:shared_warps}
\end{equation}
It immediately follows that
\begin{equation}
    \mathbf{p}+\mathbf{M}_{0\rightarrow t}(\mathbf{p})
    =\mathbf{q}+\mathbf{M}_{1\rightarrow t}(\mathbf{q})
    =\mathbf{p}_t.
    \label{eq:same_intersection}
\end{equation}
Thus, corresponding pixels transported from the two anchors must meet at the same location in every intermediate frame, ensuring consistent motion trajectories between the two anchors. Visibility weights are predicted separately to determine how much each endpoint contributes when a point is occluded or multiple points arrive at the same target location.

\paragraph{Pixel and feature warping.}
The shared trajectory is used for two forward-warping operations. First, we bicubically upsample the anchor RGB frames and forward-warp them with $\mathbf{M}_{0\rightarrow t}$ and $\mathbf{M}_{1\rightarrow t}$. Confidence-weighted soft splatting\cite{niklaus2020softmax} combines the two contributions into a coarse frame $\widetilde I_t^{H}$. This frame provides reliable color and low-frequency structure, although it may be blurry around motion boundaries or uncovered regions.

Second, we apply the same two motions to the lifted features
$\mathbf{F}_0^{H}$ and $\mathbf{F}_1^{H}$:
\begin{equation}
    \mathbf{F}_t^{H}
    =\operatorname{SoftSplat}
      (\mathbf{F}_0^{H},\mathbf{F}_1^{H};
       \mathbf{M}_{0\rightarrow t},\mathbf{M}_{1\rightarrow t}).
    \label{eq:feature_warp}
\end{equation}
Pixel warping and feature warping therefore share exactly the same geometry. The former gives an interpretable image-space estimate, while the latter brings detailed HR appearance evidence to the same target locations.

\paragraph{Residual reconstruction.}
Conditioned on $(t,s)$, a lightweight pointwise decoder combines the warped appearance and feature representation with the upsampled interval-context feature $\mathbf{F}_{01}^{H}$ to predict a bounded RGB residual. Adding this residual to $\widetilde I_t^{H}$ gives the final output in Eq.~\eqref{eq:final_output}. In this way, explicit warping handles motion and color transport, while the decoder focuses on disocclusions, splatting artifacts, and fine details.
% The decoder takes the aligned HR feature, the coarse frame, the two
% directional warped RGB frames, their contribution weights, and the query
% conditions $(t,s)$. A small pointwise network predicts a bounded RGB residual.
% Adding this residual to $\widetilde I_t^{H}$ gives the final output in
% Eq.~\eqref{eq:final_output}. In this way, explicit warping handles motion and
% color transport, while the decoder focuses on disocclusions, splatting
% artifacts, and fine details.

\subsection{Training Objective}

We train the network with four objectives. First, the final reconstruction is
supervised by a Charbonnier loss
\begin{equation}
    \mathcal{L}_{\mathrm{img}}
    =\rho(\widehat I_t^{H}-I_t^{H}),
    \qquad \rho(x)=\sqrt{x^2+\epsilon^2}.
    \label{eq:image_loss}
\end{equation}
Second, when several target times are sampled from the same interval, we
match their first-order temporal changes:
\begin{equation}
    \mathcal{L}_{\mathrm{temp}}
    =\rho\!\left(
      (\widehat I_{t_{j+1}}^{H}-\widehat I_{t_j}^{H})
      -(I_{t_{j+1}}^{H}-I_{t_j}^{H})
    \right).
    \label{eq:temporal_loss}
\end{equation}
Third, we directly supervise the motions predicted by PixelSpline using pseudo-ground-truth optical flows generated by a frozen SEA-RAFT model. 
% Third, teacher flows directly supervise the motions obtained from PixelSpline.
Let $\mathbf{M}_{i\rightarrow t}^{*}$ be the optical flow estimated by SEA-RAFT from anchor $i$ to time $t$. The flow loss is
\begin{equation}
    \mathcal{L}_{\mathrm{flow}}
    =\frac{1}{2}\sum_{i\in\{0,1\}}
      \rho\!\left(
        \frac{\mathbf{M}_{i\rightarrow t}
              -\mathbf{M}_{i\rightarrow t}^{*}}
             {(W_s,H_s)}
      \right),
    \label{eq:flow_loss}
\end{equation}
where the horizontal and vertical errors are normalized by the output width and height, respectively, so that the loss is comparable across spatial scales. Finally, the coarse warped frame is encouraged to match a blurry target obtained by downsampling and then upsampling the HR ground truth:
\begin{equation}
    \mathcal{L}_{\mathrm{warp}}
    =\rho(\widetilde I_t^{H}-I_t^{\mathrm{blur}}).
    \label{eq:warp_loss}
\end{equation}
This term supervises the explicit pixel-transport path without forcing it to reconstruct details that should be recovered by the residual decoder.

The complete objective is
\begin{equation}
    \mathcal{L}
    =\mathcal{L}_{\mathrm{img}}
     +\lambda_{\mathrm{temp}}\mathcal{L}_{\mathrm{temp}}
     +\lambda_{\mathrm{flow}}\mathcal{L}_{\mathrm{flow}}
     +\lambda_{\mathrm{warp}}\mathcal{L}_{\mathrm{warp}}.
    \label{eq:total_loss}
\end{equation}

\section{Experiments}
\begin{table*}[t]
    \centering
    \caption{Quantitative comparison of C-STVSR reconstruction quality under $\mathrm{S}\!\times\!4$ spatial and $\mathrm{T}\!\times\!8$ temporal upsampling, reported as PSNR $\uparrow$ / SSIM $\uparrow$. Best and second-best results are highlighted in bold and underlined, respectively.}
    \label{tab:quality_comparison}
    \setlength{\tabcolsep}{3.5pt}
    \resizebox{\textwidth}{!}{%
    \begin{tabular}{lcccccc}
        \toprule
        Method
        & GoPro Center
        & GoPro Average
        & Adobe Center
        & Adobe Average
        & REDS Center
        & REDS Average \\
        \midrule
        Zooming SlowMo~\cite{xiang2020zooming}
        & 30.69 / 0.885 & --
        & 30.26 / 0.882 & --
        & -- & -- \\

        TMNet~\cite{xu2021temporal}
        & 24.94 / 0.840 & 24.76 / 0.824
        & 27.25 / 0.843 & 26.65 / 0.827
        & 22.64 / \underline{0.667} & 19.84 / \textbf{0.573} \\

        VideoINR~\cite{chen2022videoinr}
        & 30.26 / 0.879 & 29.41 / 0.867
        & 29.92 / 0.875 & 29.27 / 0.865
        & 23.25 / 0.650 & 19.85 / \underline{0.558} \\

        MoTIF~\cite{chen2023motif}
        & 31.12 / 0.889 & 30.14 / 0.879
        & 30.63 / 0.884 & 29.82 / 0.875
        & 23.62 / 0.654 & 20.01 / 0.556 \\

        BF-STVSR~\cite{kim2025bf}
        & 31.22 / 0.890 & 30.26 / 0.881
        & 30.83 / 0.888 & 30.12 / 0.880
        & 23.55 / 0.646 & 19.95 / 0.549 \\

        V$^3$~\cite{becker2025continuous}
        & \textbf{32.98 / 0.923} & \textbf{32.27 / 0.919}
        & \textbf{32.91 / 0.922} & \textbf{32.29 / 0.917}
        & \textbf{24.00 / 0.671} & \textbf{20.36 / 0.573} \\

        \midrule
        \textbf{Ours}
        & \underline{32.21 / 0.912}
        & \underline{31.47 / 0.906}
        & \underline{31.99 / 0.913}
        & \underline{31.39 / 0.907}
        & \underline{23.78} / 0.651
        & \underline{20.04} / 0.549 \\
        \bottomrule
    \end{tabular}%
    }
\end{table*}

We evaluate our method on the full continuous space-time video super-resolution (C-STVSR) task, as well as on spatial-only and temporal-only reconstruction. We report reconstruction accuracy, temporal consistency, and end-to-end runtime to assess the quality achieved by each method relative to its inference cost.

\subsection{Experimental Setup}

\paragraph{Implementation and training details.}
We train our model in PyTorch on four NVIDIA H100 GPUs with a total batch size of 8. Each sample contains eight temporally subsampled context frames. For
supervision, the first and last frames of a nine-frame high-frame-rate sequence
define the endpoint pair, and three timestamps are randomly sampled from the
sequence as targets. Training uses $64\times64$ low-resolution crops with random flipping and rotation. We train the model with Adam optimizer ($\beta_1=0.9$, $\beta_2=0.99$), using an initial learning rate of $10^{-4}$ and cosine annealing to $10^{-7}$. The pretrained SEA-RAFT~\cite{wang2024sea} and GMFlow~\cite{xu2022gmflow} components in Shared PixelSpline Modeling remain frozen, while all other components are trained from scratch.

\paragraph{Datasets.}
Following the standard C-STVSR protocol~\cite{chen2022videoinr,chen2023motif,kim2025bf,becker2025continuous}, we train on the training split of Adobe240~\cite{su2017deep}. Adobe240 contains 133 high-frame-rate videos captured at 240 fps with a spatial resolution of $1280\times720$. We construct the low-resolution inputs by bicubic spatial downsampling and temporally subsample the videos by a factor of $8$, producing 30-fps input sequences. The remaining high-frame-rate frames provide supervision at continuous temporal coordinates. We report in-distribution results on the Adobe240 test split and cross-dataset results on GoPro~\cite{kim2017deep} and REDS~\cite{Nah_2019_CVPR_Workshops_REDS}. We additionally use Vid4~\cite{liu2011bayesian} to evaluate the spatial-only and temporal-only boundary cases. No dataset-specific fine-tuning is performed for GoPro, REDS, or Vid4.

\paragraph{Baseline methods.}
We compare against six representative STVSR methods. Zooming Slow-Mo~\cite{xiang2020zooming} and TMNet~\cite{xu2021temporal} are fixed-scale one-stage models, whereas VideoINR~\cite{chen2022videoinr}, MoTIF~\cite{chen2023motif}, BF-STVSR~\cite{kim2025bf}, and V$^3$~\cite{becker2025continuous} support continuous space-time reconstruction. Fixed-scale methods are evaluated only at scaling factors supported by their pretrained models; continuous methods are queried directly at the requested spatial and temporal coordinates.

\paragraph{Evaluation metrics.}
We evaluate reconstruction quality using PSNR and SSIM. In the joint C-STVSR evaluation, ``Center'' measures the input keyframes and the center frame of each interval, whereas ``Average'' includes every reconstructed output frame. The Adobe240, GoPro, and REDS experiments use $\mathrm{S}\!\times\!4$ spatial and $\mathrm{T}\!\times\!8$ temporal upsampling. For the boundary-case evaluation on Vid4, we separately set the temporal factor to $\times1$ for spatial-only SR and the spatial factor to $\times1$ for temporal-only SR. We further report temporal optical-flow consistency (tOF), end-to-end latency, and FPS.

% \begin{figure*}[t]
%     \centering
%     \includegraphics[width=\textwidth]{figures/mainresult.png}
%     \caption{Qualitative comparison of C-STVSR methods under
% $\mathrm{S}\!\times\!4$ spatial and $\mathrm{T}\!\times\!8$ temporal
% upsampling. The first three rows show results on GoPro, and the last two rows
% show results on REDS. Our method preserves fine details such as characters,
% texture patterns, and clothing structures while producing sharp and smooth
% boundaries. Best viewed zoomed in.
% }
% \label{fig:main}
%     \label{fig:main}
% \end{figure*}

\subsection{Quality--Efficiency Trade-off}

Tab.~\ref{tab:quality_comparison} compares reconstruction accuracy on the full C-STVSR task. V$^3$ achieves the highest PSNR, and our method ranks second in all six dataset--protocol combinations. The PSNR difference between the two methods is 0.77/0.80 dB on GoPro Center/Average, 0.92/0.90 dB on Adobe240 Center/Average, and 0.22/0.32 dB on REDS Center/Average. Our method also outperforms every other baseline in PSNR across all reported columns. 

Tab.~\ref{tab:inference_efficiency} reports end-to-end latency and throughput under the timing protocol described in the caption. As shown in Fig.~\ref{fig:quality_latency_tradeoff}, our method requires 406.41 ms per output frame, compared with 912.21 ms for V$^3$. This corresponds to a $2.24\times$ speedup, with a 0.90 dB lower PSNR on Adobe240 Average. Methods with lower latency than ours operate at lower reconstruction accuracy, while MoTIF and BF-STVSR are both slower and less accurate. Our method therefore provides a practical balance: it approaches the reconstruction quality of V$^3$ while requiring less than half of its end-to-end inference time.
\begin{table}[!htb]
    \centering
    \caption{End-to-end inference efficiency on Adobe240 after one full-resolution warm-up pass. Latency includes input loading and preprocessing, device transfers, all method-specific inference operations, image conversion, PNG encoding, and disk writing.}
    \label{tab:inference_efficiency}
    \small
    \setlength{\tabcolsep}{5pt}
    \begin{tabular}{lcc}
        \toprule
        Method
        & Latency (ms/frame) $\downarrow$
        & FPS $\uparrow$ \\
        \midrule
        Zooming SlowMo~\cite{xiang2020zooming}
        & 354.72 & 2.82 \\
        TMNet~\cite{xu2021temporal}
        & 81.44 & 12.28 \\
        VideoINR~\cite{chen2022videoinr}
        & 360.58 & 2.77 \\
        MoTIF~\cite{chen2023motif}
        & 494.65 & 2.02 \\
        BF-STVSR~\cite{kim2025bf}
        & 417.51 & 2.40 \\
        V$^3$~\cite{becker2025continuous}
        & 912.21 & 1.10 \\
        \midrule
        \textbf{Ours}
        & \textbf{406.41}
        & \textbf{2.46} \\
        \bottomrule
    \end{tabular}
\end{table}

\paragraph{Qualitative comparison.}
\begin{figure*}[t]
    \centering
    \includegraphics[width=\textwidth]{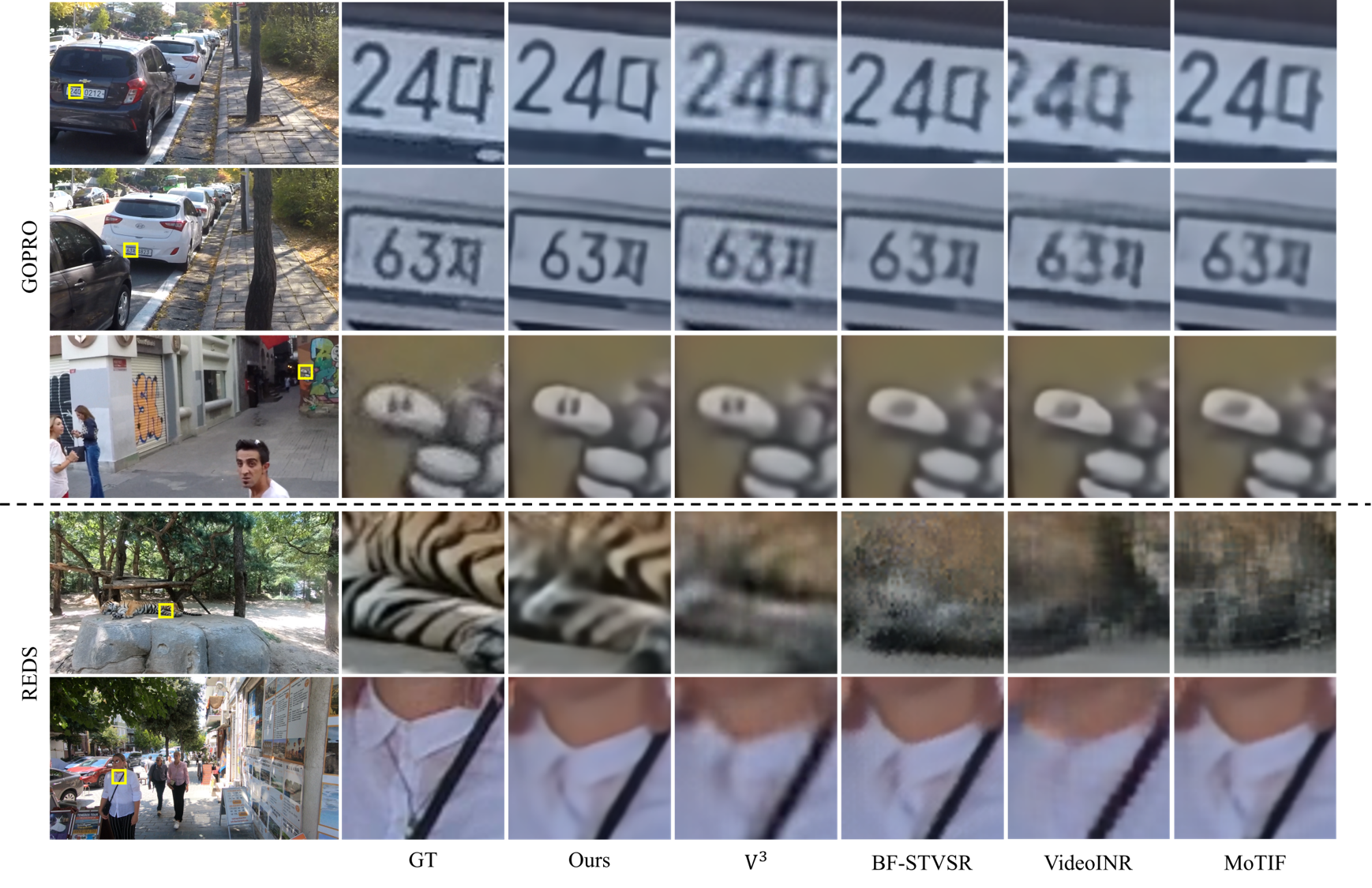}
    \caption{Qualitative comparison of C-STVSR methods under $\mathrm{S}\!\times\!4$ spatial and $\mathrm{T}\!\times\!8$ temporal upsampling. The first three rows show results on GoPro, and the last two rows show results on REDS. Our method preserves fine details such as characters, texture patterns, and clothing structures while producing sharp and smooth boundaries. Best viewed zoomed in. 
}
\label{fig:main}
    \label{fig:main}
\end{figure*}
Fig.~\ref{fig:main} presents qualitative comparisons under $\mathrm{S}\!\times\!4$ spatial and $\mathrm{T}\!\times\!8$ temporal
upsampling. The first three rows show results on GoPro, while the last two rows show results on the more challenging REDS dataset. Across both datasets, our method better preserves fine textures and high-frequency structures. In the GoPro examples, it reconstructs sharper characters, whereas V$^3$ and BF-STVSR produce noticeably blurred text. This advantage remains evident on REDS: our method faithfully recovers the fine stripe patterns on the tiger and preserves the folded structure of the pedestrian's shirt collar with smooth, well-defined boundaries. In comparison, V$^3$ and VideoINR over-smooth the collar and fail to separate its structural components, while BF-STVSR and MoTIF recover only its coarse shape and introduce jagged artifacts along the edges. These results demonstrate the ability of our method to preserve fine appearance details and coherent object boundaries across different data distributions.

\subsection{Temporal Consistency}

\begin{table}[!htb]
    \centering
    \caption{
        Temporal consistency on REDS under
        $\mathrm{S}\!\times\!4$ spatial and
        $\mathrm{T}\!\times\!8$ temporal upsampling, measured by
        tOF $\downarrow$.
    }
    \label{tab:reds_tof}
    \resizebox{\columnwidth}{!}{%
        \begin{tabular}{lcccccc}
            \toprule
            Method
            & TMNet
            & VideoINR
            & MoTIF
            & BF-STVSR
            & V$^3$
            & Ours \\
            \midrule
            tOF $\downarrow$
            & 9.944
            & 9.495
            & 9.391
            & 9.237
            & \underline{8.923}
            & \textbf{8.171} \\
            \bottomrule
        \end{tabular}%
    }
\end{table}

\begin{figure*}[t]
    \centering
    \includegraphics[width=\textwidth]{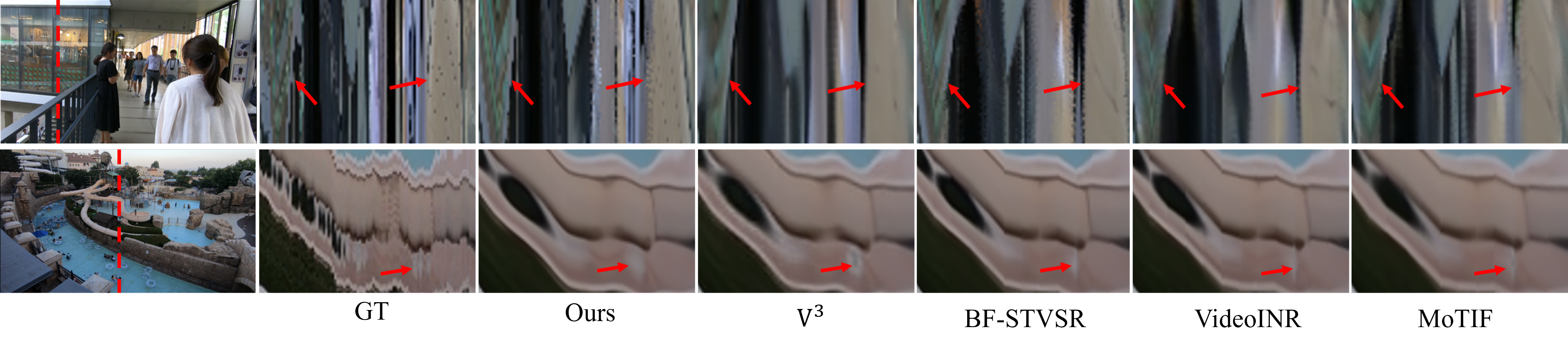}
    \caption{Qualitative comparison of temporal consistency. The red dashed line marks the vertical pixel column sampled from each frame to construct the temporal profile, while the red arrows indicate regions with noticeable deviations from the ground truth. Our method faithfully reconstructs fine details along nonlinear motion trajectories while suppressing motion-induced artifacts.
}
    \label{fig:temporal_consistancy}
\end{figure*}

Following prior C-STVSR evaluation~\cite{becker2025continuous}, we compute tOF as the endpoint error between the optical flow of adjacent reconstructed frames and the corresponding ground-truth flow. Lower tOF indicates that the reconstructed motion is closer to the reference and is therefore less affected by temporal artifacts such as flicker or irregular motion. Tab.~\ref{tab:reds_tof} shows that our method achieves the lowest tOF on REDS, reducing the error from 8.923 for V$^3$ to 8.171. Fig.~\ref{fig:temporal_consistancy} illustrates qualitative results of temporal consistency, our method successfully reconstructs temporal motion while suppressing motion artifacts.

\subsection{Decoupled Spatial and Temporal Reconstruction}

Although our model is designed for joint space--time super-resolution, its continuous formulation supports both boundary cases without architectural changes. Setting the temporal factor to $\times1$ isolates spatial super-resolution, while setting the spatial factor to $\times1$ isolates temporal interpolation.

\begin{table}[!htb]
    \centering
    \caption{Decoupled spatial-only and temporal-only super-resolution results on Vid4 (PSNR $\uparrow$ / SSIM $\uparrow$). Temporal-only results are evaluated on center frames. Best and second-best results are highlighted in bold and underlined, respectively.}
    \label{tab:vid4_axis_controls}
    \resizebox{\columnwidth}{!}{%
        \begin{tabular}{lccccc}
            \toprule
            Setting
            & VideoINR~\cite{chen2022videoinr}
            & MoTIF~\cite{chen2023motif}
            & BF-STVSR~\cite{kim2025bf}
            & V$^3$~\cite{becker2025continuous}
            & Ours \\
            \midrule

            $\mathrm{S}\!\times\!4,\,
             \mathrm{T}\!\times\!1$
            & 25.97 / 0.775
            & 26.21 / 0.780
            & 26.28 / 0.782
            & \textbf{27.15} / \textbf{0.819}
            & \underline{26.78} / \underline{0.808} \\

            $\mathrm{S}\!\times\!1,\,
             \mathrm{T}\!\times\!8$
            & 19.76 / 0.462
            & 28.86 / \underline{0.864}
            & \underline{29.42} / 0.810
            & \textbf{31.15} / 0.862
            & 29.01 / \textbf{0.890} \\

            \bottomrule
        \end{tabular}%
    }
\end{table}

Tab.~\ref{tab:vid4_axis_controls} reports both settings on Vid4. For spatial-only $\mathrm{S}\!\times\!4,\mathrm{T}\!\times\!1$ reconstruction, our method obtains 26.78 dB/0.808, ranking second behind V$^3$. This result shows that the continuous feature-lifting module remains effective when temporal synthesis is disabled. For temporal-only $\mathrm{S}\!\times\!1,\mathrm{T}\!\times\!8$ interpolation, our method achieves the highest SSIM (0.890), although its PSNR (29.01 dB) performs below V$^3$ and BF-STVSR. Overall, the two boundary cases show that the model remains competitive when spatial and temporal reconstruction are evaluated separately.

\subsection{Ablation study.}
We evaluate all variants using their checkpoints at 10,000 iterations on the same randomly selected subset of the Adobe240 test set, 
comprising 3,858 frames under $\mathrm{S}\!\times\!4$ spatial and $\mathrm{T}\!\times\!8$ temporal upsampling. The full model achieves 32.49\,dB PSNR and 0.930 SSIM. Replacing the shared bidirectional trajectory controls with independently predicted forward and backward controls, while retaining the long-context architecture, reduces performance to 32.25\,dB and 0.924, corresponding to a 0.24\,dB PSNR drop. Removing the additional context frames and using only the two anchor frames further decreases performance to 31.80\,dB and 0.918, a 0.69\,dB drop relative to the full model. These results verify that both endpoint-consistent control sharing and long-range temporal context contribute to reconstruction quality.
\section{Conclusion}

We presented a continuous space-time video super-resolution framework that combines long-context encoding, continuous HR feature upsampling, and an endpoint-anchored PixelSpline. By constructing a single correspondence-consistent trajectory shared by both endpoints, our method aligns pixel and feature transport while exploiting observations beyond the queried frame pair. Experiments across joint, spatial-only, and temporal-only super-resolution settings demonstrate competitive reconstruction quality, improved temporal consistency, and a favorable quality--efficiency trade-off.

Despite these results, the reconstruction performance in terms of PSNR and SSIM can be further improved, particularly compared with the quality-leading method in several settings. Moreover, although our method offers practical inference efficiency, its training process remains computationally demanding and time-consuming. Improving reconstruction accuracy while developing more efficient training strategies constitutes a promising direction for future work.

\section{Acknowledgment}
This work was supported in part by the U.S. National Science Foundation (NSF) under Awards IIS-2232298, IIS-2442792, and OIA-2242812.

{
    \small
    \bibliographystyle{ieeenat_fullname}
    \bibliography{main}

@String(CVPR= {IEEE Conf. Comput. Vis. Pattern Recog.})

@String(ICCV= {Int. Conf. Comput. Vis.})

@String(AAAI = {AAAI})

@String(CVPRW= {IEEE Conf. Comput. Vis. Pattern Recog. Worksh.})

@String(CVPR  = {CVPR})

@String(ICCV  = {ICCV})

@String(CVPRW= {CVPRW})

@inproceedings{jiang2018super,
  title={Super slomo: High quality estimation of multiple intermediate frames for video interpolation},
  author={Jiang, Huaizu and Sun, Deqing and Jampani, Varun and Yang, Ming-Hsuan and Learned-Miller, Erik and Kautz, Jan},
  booktitle={Proceedings of the IEEE conference on computer vision and pattern recognition},
  pages={9000--9008},
  year={2018}
}

@inproceedings{bao2019depth,
  title={Depth-aware video frame interpolation},
  author={Bao, Wenbo and Lai, Wei-Sheng and Ma, Chao and Zhang, Xiaoyun and Gao, Zhiyong and Yang, Ming-Hsuan},
  booktitle={2019 IEEE/CVF Conference on Computer Vision and Pattern Recognition (CVPR)},
  pages={3698--3707},
  year={2019},
  organization={IEEE}
}

@inproceedings{zhang2018image,
  title={Image super-resolution using very deep residual channel attention networks},
  author={Zhang, Yulun and Li, Kunpeng and Li, Kai and Wang, Lichen and Zhong, Bineng and Fu, Yun},
  booktitle={European conference on computer vision},
  pages={294--310},
  year={2018},
  organization={Springer}
}

@inproceedings{wang2019edvr,
  title={Edvr: Video restoration with enhanced deformable convolutional networks},
  author={Wang, Xintao and Chan, Kelvin CK and Yu, Ke and Dong, Chao and Loy, Chen Change},
  booktitle={2019 IEEE/CVF Conference on Computer Vision and Pattern Recognition Workshops (CVPRW)},
  pages={1954--1963},
  year={2019},
  organization={IEEE}
}

@inproceedings{xiang2020zooming,
  title={Zooming slow-mo: Fast and accurate one-stage space-time video super-resolution},
  author={Xiang, Xiaoyu and Tian, Yapeng and Zhang, Yulun and Fu, Yun and Allebach, Jan P and Xu, Chenliang},
  booktitle={2020 IEEE/CVF Conference on Computer Vision and Pattern Recognition (CVPR)},
  pages={3367--3376},
  year={2020},
  organization={IEEE}
}

@article{xu2021temporal,
  title={Temporal modulation network for controllable space-time video super-resolution},
  author={Xu, Gang and Xu, Jun and Li, Zhen and Wang, Liang and Sun, Xing and Cheng, Ming-Ming},
  journal={arXiv preprint arXiv:2104.10642},
  year={2021}
}

@inproceedings{geng2022rstt,
  title={Rstt: Real-time spatial temporal transformer for space-time video super-resolution},
  author={Geng, Zhicheng and Liang, Luming and Ding, Tianyu and Zharkov, Ilya},
  booktitle={2022 IEEE/CVF Conference on Computer Vision and Pattern Recognition (CVPR)},
  pages={17420--17430},
  year={2022},
  organization={IEEE}
}

@inproceedings{haris2020space,
  title={Space-time-aware multi-resolution video enhancement},
  author={Haris, Muhammad and Shakhnarovich, Greg and Ukita, Norimichi},
  booktitle={2020 IEEE/CVF Conference on Computer Vision and Pattern Recognition (CVPR)},
  pages={2856--2865},
  year={2020},
  organization={IEEE}
}

@inproceedings{hu2022you,
  title={You only align once: Bidirectional interaction for spatial-temporal video super-resolution},
  author={Hu, Mengshun and Jiang, Kui and Nie, Zhixiang and Wang, Zheng},
  booktitle={Proceedings of the 30th ACM International Conference on Multimedia},
  pages={847--855},
  year={2022}
}

@inproceedings{hu2023store,
  title={Store and fetch immediately: Everything is all you need for space-time video super-resolution},
  author={Hu, Mengshun and Jiang, Kui and Nie, Zhixiang and Zhou, Jiahuan and Wang, Zheng},
  booktitle={Proceedings of the AAAI Conference on Artificial Intelligence},
  volume={37},
  number={1},
  pages={863--871},
  year={2023}
}

@inproceedings{choi2026time,
  title={Time Without Time: Pseudo-Temporal Representation for Space-Time Super-Resolution},
  author={Choi, Hee Min and Kang, Hyoa and Kim, Suji and Oh, Dokwan and Cho, Nam Ik},
  booktitle={Proceedings of the IEEE/CVF Conference on Computer Vision and Pattern Recognition},
  pages={6812--6822},
  year={2026}
}

@inproceedings{hu2019meta,
  title={Meta-SR: A magnification-arbitrary network for super-resolution},
  author={Hu, Xuecai and Mu, Haoyuan and Zhang, Xiangyu and Wang, Zilei and Tan, Tieniu and Sun, Jian},
  booktitle={2019 IEEE/CVF Conference on Computer Vision and Pattern Recognition (CVPR)},
  pages={1575--1584},
  year={2019},
  organization={IEEE}
}

@inproceedings{chen2021learning,
  title={Learning continuous image representation with local implicit image function},
  author={Chen, Yinbo and Liu, Sifei and Wang, Xiaolong},
  booktitle={Proceedings of the IEEE/CVF conference on computer vision and pattern recognition},
  pages={8628--8638},
  year={2021}
}

@inproceedings{lee2022local,
  title={Local texture estimator for implicit representation function},
  author={Lee, Jaewon and Jin, Kyong Hwan},
  booktitle={2022 IEEE/CVF Conference on Computer Vision and Pattern Recognition (CVPR)},
  pages={1919--1928},
  year={2022},
  organization={IEEE}
}

@article{becker2023thera,
  title={Thera: Aliasing-free arbitrary-scale super-resolution with neural heat fields},
  author={Becker, Alexander and Daudt, Rodrigo Caye and Narnhofer, Dominik and Peters, Torben and Metzger, Nando and Wegner, Jan Dirk and Schindler, Konrad},
  journal={arXiv preprint arXiv:2311.17643},
  year={2023}
}

@inproceedings{li2024savsr,
  title={SAVSR: Arbitrary-scale video super-resolution via a learned scale-adaptive network},
  author={Li, Zekun and Liu, Hongying and Shang, Fanhua and Liu, Yuanyuan and Wan, Liang and Feng, Wei},
  booktitle={Proceedings of the AAAI Conference on Artificial Intelligence},
  volume={38},
  number={4},
  pages={3288--3296},
  year={2024}
}

@inproceedings{shang2024arbitrary,
  title={Arbitrary-scale video super-resolution with structural and textural priors},
  author={Shang, Wei and Ren, Dongwei and Zhang, Wanying and Fang, Yuming and Zuo, Wangmeng and Ma, Kede},
  booktitle={European Conference on Computer Vision},
  pages={73--90},
  year={2024},
  organization={Springer}
}

@inproceedings{chen2022videoinr,
  title={Videoinr: Learning video implicit neural representation for continuous space-time super-resolution},
  author={Chen, Zeyuan and Chen, Yinbo and Liu, Jingwen and Xu, Xingqian and Goel, Vidit and Wang, Zhangyang and Shi, Humphrey and Wang, Xiaolong},
  booktitle={2022 IEEE/CVF Conference on Computer Vision and Pattern Recognition (CVPR)},
  pages={2037--2047},
  year={2022},
  organization={IEEE}
}

@inproceedings{chen2023motif,
  title={Motif: Learning motion trajectories with local implicit neural functions for continuous space-time video super-resolution},
  author={Chen, Yi-Hsin and Chen, Si-Cun and Lin, Yen-Yu and Peng, Wen-Hsiao},
  booktitle={2023 IEEE/CVF International Conference on Computer Vision (ICCV)},
  pages={23074--23084},
  year={2023},
  organization={IEEE}
}

@inproceedings{kim2025bf,
  title={Bf-stvsr: B-splines and fourier—best friends for high fidelity spatial-temporal video super-resolution},
  author={Kim, Eunjin and Kim, Hyeonjin and Jin, Kyong Hwan and Yoo, Jaejun},
  booktitle={2025 IEEE/CVF Conference on Computer Vision and Pattern Recognition (CVPR)},
  pages={28009--28018},
  year={2025},
  organization={IEEE}
}

@article{becker2025continuous,
  title={Continuous space-time video super-resolution with 3d fourier fields},
  author={Becker, Alexander and Erbach, Julius and Narnhofer, Dominik and Schindler, Konrad},
  journal={arXiv preprint arXiv:2509.26325},
  year={2025}
}

@inproceedings{liu2024sparse,
  title={Sparse global matching for video frame interpolation with large motion},
  author={Liu, Chunxu and Zhang, Guozhen and Zhao, Rui and Wang, Limin},
  booktitle={2024 IEEE/CVF Conference on Computer Vision and Pattern Recognition (CVPR)},
  pages={19125--19134},
  year={2024},
  organization={IEEE}
}

@inproceedings{wang2024sea,
  title={Sea-raft: Simple, efficient, accurate raft for optical flow},
  author={Wang, Yihan and Lipson, Lahav and Deng, Jia},
  booktitle={European Conference on Computer Vision},
  pages={36--54},
  year={2024},
  organization={Springer}
}

@inproceedings{xu2022gmflow,
  title={Gmflow: Learning optical flow via global matching},
  author={Xu, Haofei and Zhang, Jing and Cai, Jianfei and Rezatofighi, Hamid and Tao, Dacheng},
  booktitle={2022 IEEE/CVF Conference on Computer Vision and Pattern Recognition (CVPR)},
  pages={8111--8120},
  year={2022},
  organization={IEEE}
}

@inproceedings{su2017deep,
  title={Deep video deblurring for hand-held cameras},
  author={Su, Shuochen and Delbracio, Mauricio and Wang, Jue and Sapiro, Guillermo and Heidrich, Wolfgang and Wang, Oliver},
  booktitle={2017 IEEE Conference on computer vision and pattern recognition (CVPR)},
  pages={237--246},
  year={2017},
  organization={IEEE}
}

@inproceedings{kim2017deep,
  title={Deep multi-scale convolutional neural network for dynamic scene deblurring},
  author={Kim, Tae Hyun and Nah, Seungjun and Lee, Kyoung Mu},
  booktitle={Conference on Computer Vision and Pattern Recognition},
  pages={1--21},
  year={2017},
  organization={IEEE}
}

@inproceedings{liu2011bayesian,
  title={A bayesian approach to adaptive video super resolution},
  author={Liu, Ce and Sun, Deqing},
  booktitle={CVPR 2011},
  pages={209--216},
  year={2011},
  organization={IEEE}
}

@InProceedings{Nah_2019_CVPR_Workshops_REDS,
  author = {Nah, Seungjun and Baik, Sungyong and Hong, Seokil and Moon, Gyeongsik and Son, Sanghyun and Timofte, Radu and Lee, Kyoung Mu},
  title = {NTIRE 2019 Challenge on Video Deblurring and Super-Resolution: Dataset and Study},
  booktitle = {CVPR Workshops},
  month = {June},
  year = {2019}
}

@inproceedings{niklaus2020softmax,
  title={Softmax splatting for video frame interpolation},
  author={Niklaus, Simon and Liu, Feng},
  booktitle={2020 IEEE/CVF Conference on Computer Vision and Pattern Recognition (CVPR)},
  pages={5436--5445},
  year={2020},
  organization={IEEE}
}

@article{sitzmann2020implicit,
  title={Implicit neural representations with periodic activation functions},
  author={Sitzmann, Vincent and Martel, Julien and Bergman, Alexander and Lindell, David and Wetzstein, Gordon},
  journal={Advances in neural information processing systems},
  volume={33},
  pages={7462--7473},
  year={2020}
}
}

\end{document}